\pdfoutput=1

\documentclass[11pt]{article}

\usepackage{acl}

\usepackage{times}
\usepackage{latexsym}
\usepackage{enumitem}
\usepackage{graphicx}
\usepackage{url}
\usepackage{amsmath}
\usepackage{cases}
\usepackage{amssymb}
\usepackage{amsthm}
\usepackage{breqn}
\usepackage{algorithm}
\usepackage{algpseudocode}
\usepackage{enumitem}
\usepackage{caption}
\usepackage{subcaption}
\usepackage{titlesec}
\usepackage{pgfplots}
\usepackage{float}
\usepackage{hyperref}

\titleformat{\subsubsection}[runin]{\bfseries}{}{}{}[]
\theoremstyle{definition}
\newtheorem{definition}{Definition}[section]
\newtheorem{theorem}{Theorem}[section]

\usepackage[T1]{fontenc}

\usepackage[utf8]{inputenc}

\usepackage{microtype}

\usepackage{inconsolata}

\title{Differentially Private Knowledge Distillation via Synthetic Text Generation}

\author{James Flemings \quad Murali Annavaram \\
         University of Southern California \\
         \texttt{\{jamesf17, annavara\}@usc.edu}
         }

\pgfplotsset{compat=1.18}
\begin{document}
\maketitle

\begin{abstract}
Large Language models (LLMs) are achieving state-of-the-art performance in many different downstream tasks. However, the increasing urgency of data privacy puts pressure on practitioners to train LLMs with Differential Privacy (DP) on private data. Concurrently, the exponential growth in parameter size of LLMs necessitates model compression before deployment of LLMs on resource-constrained devices or latency-sensitive applications. Differential privacy and model compression generally must trade off utility loss to achieve their objectives. Moreover, simultaneously applying both schemes can compound the utility degradation. To this end, we propose DistilDP: a novel differentially private knowledge distillation algorithm that exploits synthetic data generated by a differentially private teacher LLM. The knowledge of a teacher LLM is transferred onto the student in two ways: one way from the synthetic data itself-- the hard labels, and the other way by the output distribution of the teacher evaluated on the synthetic data-- the soft labels. Furthermore, if the teacher and student share a similar architectural structure, we can further distill knowledge by aligning the hidden representations between both. Our experimental results demonstrate that DistilDP can substantially improve the utility over existing baselines, at least $9.0$ PPL on the Big Patent dataset, with strong privacy parameters, $\epsilon=2$. These promising results progress privacy-preserving compression of autoregressive LLMs. Our code can be accessed here: \url{https://github.com/james-flemings/dp_compress}.
\end{abstract}
\section{Introduction}
Large Language Models (LLMs) have accomplished amazing feats, from generating human-like text to improving human productivity through AI-powered assistants. However, recent work has demonstrated that practical privacy attacks are possible due to memorization of training data from LLMs \cite{carlini2019secret, carlini2021extracting}. Differential Privacy (DP), a rigorous mathematical framework, is widely used to mitigate this type of information leakage \cite{dwork2006differential}. To guarantee DP for machine learning models, DP is operationalized within the learning algorithm, known as DP-SGD \cite{abadi2016deep}. The standard training paradigm of LLMs is to first pre-train an LLM on large publicly available data, then fine-tune it on a specific downstream task using private data. Recent works in differentially private training have taken advantage of this paradigm by initializing an LLM using pre-trained weights and then applying DP-SGD in the fine-tuning process, which can significantly boost its performance \cite{li2021large, yu2021differentially, ganesh2023public}, albeit with some reduction in model utility. 

Due to the huge computational and memory demands of LLMs, it becomes necessary to consider model compression in this training paradigm. For example, model compression is required when deploying LLMs on client devices where memory resources are scarce, or real-time systems such as sentence completion in text editors where inference latency is crucial \cite{xu2023federated}. One popular technique for model compression that our work focuses on is task-specific Knowledge Distillation (KD) \cite{hinton2015distilling}, which distills the knowledge of a larger teacher model onto a compressed student model. Although knowledge distillation in LLMs has been quite successful \cite{jiao2019tinybert, sanh2019distilbert, sun2019patient}, few works have explored compressing LLMs with DP \cite{mireshghallah2022differentially, yu2023training, yu2023selective}. Since achieving DP or KD for an LLM already trade-offs utility loss, accomplishing both can result in substantial utility loss, which highlights the challenge of this problem.

\begin{figure*}[t!]
    \centering
    \includegraphics[width=1.75\columnwidth]{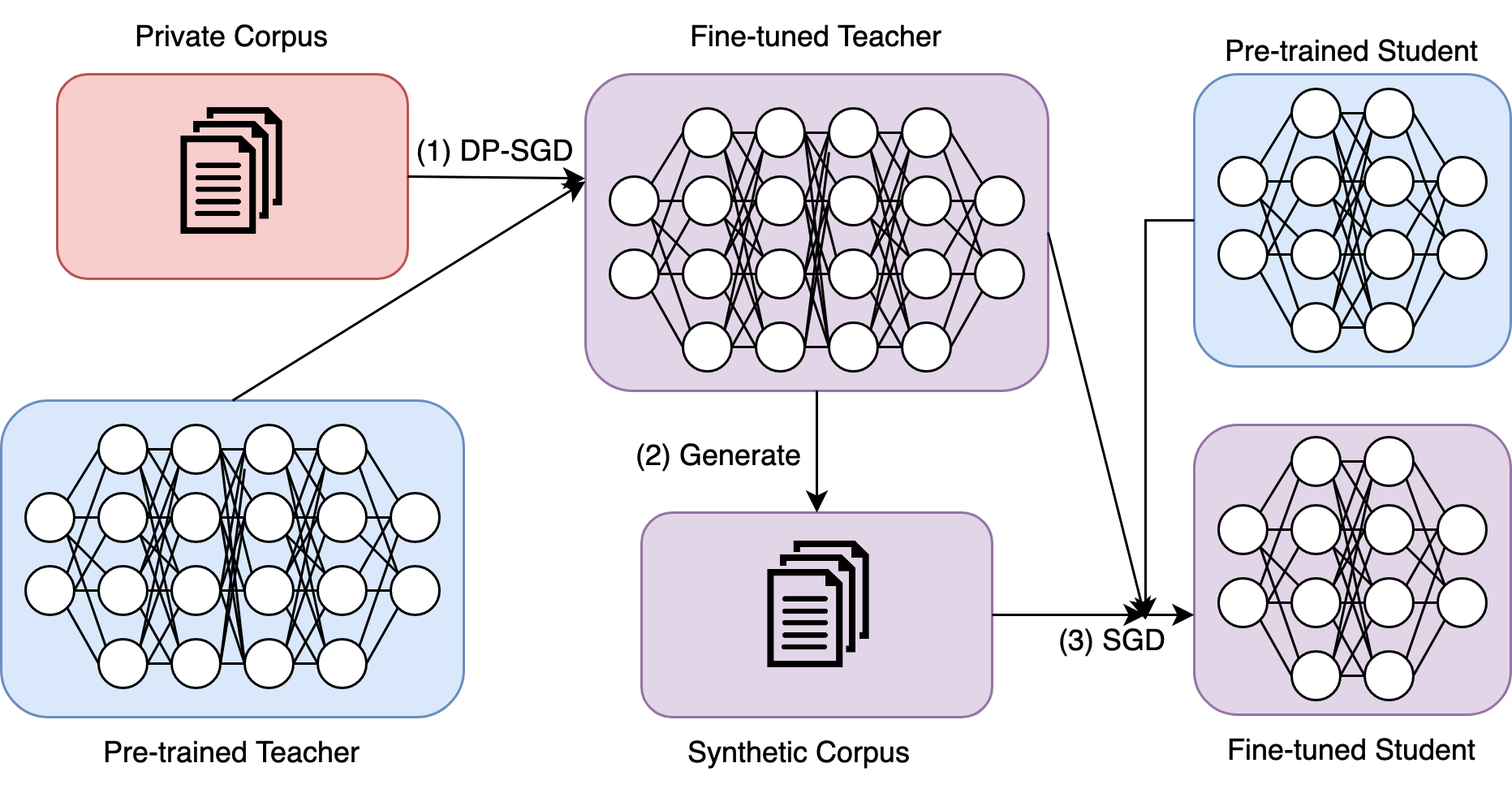}
    \caption{A brief overview of our differentially private knowledge distillation framework, DistilDP, which can be broken down in three steps: (1) A pre-trained teacher model is fine-tuned on a private corpus using DP-SGD. (2) Next, the teacher model will generate synthetic data using control codes. (3) Finally, a pre-trained student model is fine-tuned on the DP synthetic data with the DP fine-tuned teacher for knowledge distillation.}
    \label{fig:dpkd_overview}
\end{figure*}

A naive solution to this problem would be to apply DP-SGD when training the teacher model, then apply DP-SGD again during the knowledge distillation between the teacher and student. Since the teacher model is a function of the entire private corpus, applying DP-SGD during the fine-tuning of the teacher is seemingly unavoidable. Despite the additive computational and memory costs due to an additional run of DP-SGD, this solution has also been shown to have worst utility compared to differentially private fine-tuning the student without knowledge distillation \cite{mireshghallah2022differentially}. 

The goal of our work is to enable efficient differentially private compression of LLM deployment while minimizing model utility degradation. To this end, we introduce DistilDP: a simple, but effective, Differentially Private Knowledge Distillation framework, shown in Figure \ref{fig:dpkd_overview}. The innovation of our approach is that it avoids running an additional DP-SGD during the knowledge distillation by utilizing DP synthetic data. Our frameworks proceeds as follows: (1) First, DP-SGD is employed to fine-tune a teacher model on a training dataset, where the teacher accurately captures the distributional properties of the training data. (2) Then we generate synthetic text closely resembling the original data distribution without incurring additional privacy loss. (3) Lastly, the student model is trained using a non-private optimizer on the differentially private synthetic data while the teacher distills its knowledge to the student. 

In summary, our key contributions are the following: (1) We introduce DistilDP, a novel differentially private knowledge distillation algorithm for autoregressive LLMs using synthetic text generation. Our framework utilizes two forms of knowledge transfer from the teacher: one from the synthetic text data, and the other from aligning the teacher output distribution with the student. (2) We comprehensively demonstrate that our algorithm significantly improves the utility of the student model over existing baselines with strong privacy parameters, $\epsilon=2$; In particular, the alignment step between the output distribution of the teacher and student is a crucial component to boosting the performance of the student. (3) We perform extensive experiments exploring different hyperparameters to provide insight into our approach. Additionally, if the teacher and student share a similar architectural structure, we can align the hidden representation of the teacher and student to further boost the utility of our framework. 
\section{Related Works}

\textbf{Knowledge Distillation and DP Compression.} Work in knowledge distillation for LLMs typically trains a smaller student model to match its output logits with the output logits of a larger, teacher model. More advanced works align the hidden state vectors between the teacher and the student \cite{jiao2019tinybert, sanh2019distilbert, sun2019patient}. 

DP-SGD \cite{abadi2016deep} is the most widely used technique to guarantee DP for LLMs. Specifically, a pre-trained LLM trained on large amounts of public text data scraped from the internet is further fine-tuned on a private corpus $D$ using DP-SGD \cite{mcmahan2017learning, yu2021differentially, li2021large}. An orthogonal approach investigated differentially private decoding of LLMs \cite{flemings2024differentially, majmudar2022differentially, ginart2022submix}, thereby avoiding the use of DP-SGD.

\citet{mireshghallah2022differentially} first introduced DP model compression and gave a framework for DP knowledge distillation. However, their framework requires using DP-SGD twice: once for the teacher and once for the student. This procedure leads to suboptimal results for text classification tasks. \citet{wang2019private} also explored private model compression, but is limited by the number of queries to the teacher for knowledge distillation. \citet{yu2023selective} explores DP model compression during the pre-training stage; our work focuses solely on compression during fine-tuning. Another line of work similar to our approach is Data-Free Distillation \cite{chen2019data, haroush2020knowledge} where a generative model synthesizes data to substitute training data. Since no private data is used, this results in no privacy loss, $\epsilon=0$. Our approach uses a private corpus to generate synthetic data in a differentially private manner.

Another problem setup broadly related to DP knowledge distillation is private ensemble learning via PATE \cite{papernot2016semi, papernot2018scalable}. An ensemble of teacher models is trained on disjoint private data, then produces noisy labels on unlabeled public data, which comes from a similar distribution as the private data, for the student to train on. \citet{lyu2020differentially} used PATE to achieve differentially private knowledge distillation for mobile analytics. Our method does not require unlabeled public data. \newline

\noindent\textbf{DP Synthetic Data for Downstream Task.}
Promising progress has been made in generating high-fidelity, DP synthetic text \cite{bommasani2019towards, mattern2022differentially, kurakin2023harnessing, yue2022synthetic}. These works have substantially reduced the utility gap between DP-SGD on real data and SGD on DP synthetic data for classification tasks. 

\citet{yu2023training} investigated synthetic data generation from DP foundational models to train lightweight downstream models. However, our method differs from theirs by the following: (1) Their framework only considers knowledge transfer from the synthetic data. Our results demonstrate that other forms of knowledge distillation, in particular, the output distribution and hidden representations, are crucial to obtain strong performance for compressed models. (2) Their framework needs large foundational models to generate high-enough quality synthetic text data, which requires model parallelism on four V100 GPUs for fine-tuning even with LoRA \cite{hu2021lora, yu2021differentially}. And to the best of our knowledge, there is no open-source implementation of DP model parallel training. In contrast, our framework can obtain strong results without relying on large foundational models and thus does not need model parallelism. 

\section{Preliminaries}
\subsection{Knowledge Distillation}
Knowledge Distillation (KD) aims at transferring the knowledge of a teacher LLM $T$ to a smaller student LLM $S$. The standard approach is to train the student to mimic the output distribution of the teacher and the true labels. Let $f_{T}(x)$ and $f_{S}(x)$ be the logit outputs of the teacher and student networks evaluated on an input data $x$, respectively, and $\mathbf{y}$ be the true labels. The student is trained to minimize the linear combination of the supervised learning loss and the distillation loss:
\begin{align}\label{eq:kd}
    \mathcal{L}_{\text{KD}}(T, S, D) = \sum_{(x, \textbf{y})\in D}(1-\lambda) \mathcal{L}_{\text{CE}}(\mathbf{y}, \sigma(f_{S}(x)))\nonumber \\
    + \lambda t^2 \mathcal{L}_{\text{KL}}(\log(\sigma(f_{T}(x) / t)), \log \left (\sigma(f_{S}(x) / t) \right ))
\end{align}
where $\lambda$ is a weight hyperparameter, $\mathcal{L}_{\text{CE}}$ is the cross-entropy loss, $\mathcal{L}_{\text{KL}}$ is the Kullback-Leibler divergence, $\sigma$ is the softmax function, and the second term is the distillation loss which follows the softmax-temperature using a temperature parameter $t$ \cite{hinton2015distilling}. At inference time, $t$ is set to 1 to recover the standard softmax. $\textbf{y}$ is known as the hard label because of its one-hot encoding while $\sigma(f_{T}(x) / t)$ is the soft label.

\subsection{Text Generation}
Given some context vector $\mathbf{x}_t = x_1, x_2, ..., x_t$, which is a string of tokens from some vocabulary $V$, i.e. $x_i \in V $ for all $i=1, ..., t$, text generation involves sampling the next token $x_{t+1}$ using a generative language model $p$. More precisely, the output of a language model $p$ for a given context $\mathbf{x}_t$ is a likelihood function of all possible tokens $p(x_{t+1} = w | \mathbf{x}_t)$, and choosing the next token involves sampling from this probability mass function to obtain a token $\hat{x}_{t+1} \sim p(x_{t+1}| \mathbf{x}_t)$.

\subsection{Differential Privacy}
Differential Privacy (DP) is a strong privacy notion that upper bounds the amount of influence an individual in a private dataset has on the output of an algorithm. In the context of training machine learning models, this means reducing memorization of each individual and forcing the model to learn features that are shared amongst the population. We formalize this below:

\begin{definition}[Differential Privacy \cite{dwork2006differential}]
Let $\epsilon > 0$ and $\delta\in [0, 1]$. A randomized algorithm $A: \mathcal{D} \rightarrow \mathcal{R}$ satisfies $(\epsilon, \delta)$-DP if for any pair of adjacent datasets $D, D' \in \mathcal{D}$ that differ exactly in a single data sample, and for all sets $R \subseteq \mathcal{R}$ it holds that 
\begin{equation*}
    \Pr[A(D) \in R]  \leq e^{\epsilon} \Pr[A(D')\in R] + \delta.
\end{equation*}
The privacy parameters $\epsilon, \delta$ can be interpreted as follows: $\epsilon$ upper bounds the privacy loss, and $\delta$ is the probability that this guarantee does not hold. One notable property of DP that we crucially rely on in this work involves the post-processing theorem. This property ensures that for any $(\epsilon, \delta)$-DP algorithm $A$, the composition $F \circ A$ is $(\epsilon, \delta)$-DP for any deterministic or randomized function $F$. In other words, a privatized quantity cannot be un-privatized if the data is not used again.

The standard approach to integrating DP for deep learning models is by modifying the stochastic gradient descent algorithm (SGD), called DP-SGD \cite{abadi2016deep}. The modifications involve calculating per-sample gradients, rather than per-batch gradients, clipping them if their l-2 norm exceeds some threshold, then adding noise sampled from a Gaussian distribution. Note that per-sample gradient calculations typically reduce the parallel efficiency of GPUs; hence, applying DP-SGD has a negative impact on the training speed \cite{yousefpour2021opacus}. In the following section, we will make use of DP-SGD.
    
\end{definition}
\section{Method}
\subsection{Problem Formulation}

Given a task-specific private dataset $D$, a pre-trained, generative teacher model $T$, a pre-trained, generative student model $S$, and a privacy budget $(\epsilon, \delta)$, our goal is to produce a student model such that $S$ minimizes Equation \ref{eq:kd}, the size of $S$ is significantly smaller than $T$, and $S$ is $(\epsilon, \delta)$-DP with respect to $D$. We will also assume access to a set of control codes $C$ \cite{keskar2019ctrl} in our solution.

The DPKD framework introduced in \citet{mireshghallah2022differentially} tackles this problem setup by first fine-tuning a teacher model with DP-SGD, and then using DP-SGD a second time to minimize Eq \ref{eq:kd} for the student. This solution has three limitations: (1) using DP-SGD on the student model results in additional error proportional to the total number of parameters of the student model, in the worst case \cite{li2022does}; (2) allocating the privacy budget between the teacher and student can make it difficult for the student to obtain useful signal for strict privacy settings, smaller $\epsilon$ values; (3) larger runtime and memory consumption with an additional use of DP-SGD. 

Rather than applying DP-SGD twice, we take the view that only applying DP-SGD on the teacher model leads to a more optimal solution. The framework from \cite{yu2023training} follows this approach, but, as our results will show, the inclusion of more information about the teacher, such as the output distribution, substantially improves the utility of the student. 

\begin{algorithm}[tb]
    \caption{DistilDP: Differentially Private Knowledge Distillation via Synthetic Data} 
    \label{alg:dpkd_framework}
    \textbf{Input:} A pre-trained teacher model $T$, pre-trained student model $S$, private data $D$, control codes $C$, max length of synthetic examples $T_{\text{max}}$, total number of synthetic examples $N$, privacy budget $(\epsilon, \delta)$, mixing parameter $\lambda$, temperature parameter $t$ \\
    \textbf{Output:} An $(\epsilon, \delta)$-DP student model $S$
    \begin{algorithmic}[1]
        \State Preprocessing Step: Prepend control codes $C$ onto the dataset $D$, i.e., $D'[i] = C[i] + D[i]$ $\forall i\in[|D|]$.
        \State Fine-tune the teacher $T$ on the preprocessed data $D'$ using DP-SGD with privacy budget $(\epsilon, \delta)$-DP
        \State Subsample a subset of the control codes $\Tilde{C} \subseteq C$ where $|\Tilde{C}| = N$
        \For {$c \in \Tilde{C}$}
            \State Synthetic example $\Tilde{d} \sim \prod\limits_{t=1}^{T_{\text{max}}}T(x_t | x_{1:t}, c)$
            \State $\Tilde{D} = \Tilde{D} \cup \Tilde{d}$
        \EndFor
        \State Train Student $S = \min\limits_{S'} \mathcal{L}_{\text{KD}}(T, S', \Tilde{D})$
        \State \Return $S$
    \end{algorithmic}
\end{algorithm}

\subsection{DistilDP: Knowledge Distillation with DP Synthetic Text Data}
We now introduce DistilDP, described in Algorithm \ref{alg:dpkd_framework}. DistilDP fine-tunes a pre-trained, generative LLM using DP-SGD as our teacher model $T$, i.e., $T$ is $(\epsilon, \delta)$-DP. Afterward, $T$ is used to generate synthetic text data. In practice, we can use any synthetic text generation methodology, but we choose to follow an approach from \citet{yue2022synthetic}. In particular, we will utilize control codes \cite{keskar2019ctrl} to provide more explicit control over text generation. Hence, sampling the next token $\hat{x}_{t+1}$ involves conditioning the model with a control code $c$ from a set of control codes $C$, in addition with the context vector $\mathbf{x}_t$: $\hat{x}_{t+1} \sim p(x_{t+1}| \mathbf{x}_t, c)$.  

Control codes guide the model to focus on specific topics or features during text generation. For example, when generating reviews, we can prompt the model with additional information about the product type and review score. A potential control code could look like the following: "Product type: Toy; Review score: 3.0." Models learn to use these control codes in their text generations when the codes are included in the training dataset. Using our working example, the model should be equipped to generate reviews for a Toy with a review score of 3.0 after fine-tuning. For our use case, we prepend each training sample with a control code containing categorical information about the sample. Then after training, we generate text data using only the control codes as the initial context vector from the training dataset, preserving the control code distribution.

Once the synthetic data $\Tilde{D}$ has been generated, we fine-tune the pre-trained student model $S$ on $\Tilde{D}$ along with knowledge distilled from the teacher model $T$ to minimize Equation \ref{eq:kd} without DP-SGD. With this setup, we can capture the knowledge of the teacher, which is a function of the entire private dataset, in two ways: the synthetically generated data and the output distribution of the teacher. Synthetic data provides a coarse way to transfer the knowledge of a teacher model, since one-hot encodings only give one piece of information about the output of the teacher. However, the knowledge distilled from the entire output distribution of the teacher is richer, providing additional information such as prediction uncertainty of the teacher that the student can learn from. Since the synthetic data closely resembles the original private data, minimizing Equation \ref{eq:kd} using $\Tilde{D}$ gives an accurate approximation to Equation \ref{eq:kd} using $D$.

Another advantage of DistilDP is that the synthetic text generator and teacher model are the same. Hence, we do not need to split the privacy budget and instead can allocate it entirely to privately fine-tuning the teacher model. This allows for better performance for the teacher, resulting in higher-quality text generation and richer information from the output distribution. Furthermore, since our method only requires DP-SGD once, which is for the teacher, we further reduce additional memory and computational costs compared to DPKD. 

Lastly, we designed our framework to be model agnostic, making no architectural assumptions between the teacher and student. However, in section \ref{sec:hid_rep}, we show that our framework can also be improved by loosening this assumption. More specifically, we can add loss terms to Eq. \ref{eq:kd} that involve aligning the hidden representations of the teacher and student model. Our method is precisely summarized in Algorithm \ref{alg:dpkd_framework}.

\subsection{Privacy Analysis}
We will now provide a privacy analysis of our framework, which we state and prove below.

\begin{theorem}
    The output of Algorithm \ref{alg:dpkd_framework} is $(\epsilon, \delta)$-DP with respect to $D$.
\end{theorem}

\textit{Proof.} Our goal is for DistilDP to guarantee $(\epsilon, \delta)$-DP for $D$, which composes of text content for each data sample. After step 1, our method produces a teacher model $T$ that is $(\epsilon, \delta)$-DP. In step 2, sampling from $T$ incurs no extra privacy loss due to the post-processing property. However, one subtle, but crucial, detail is the potential privacy leakage from the control codes $C$. We limit the control codes to only contain categorical information of a data sample, not text contents. However, the categorical distribution in the original dataset may be private information. In our work, we follow the same procedure as \citet{yue2022synthetic} by ignoring this low-cost privacy loss and using the exact categorical distribution. For step 3, since $\Tilde{D}$ and $T$ are $(\epsilon, \delta)$-DP with respect to $D$, utilizing a non-private optimizer to minimize Equation \ref{eq:kd} will result in $S$ being $(\epsilon, \delta)$-DP due to the post-processing property. \qed

\begin{table*}[ht!]
\centering
\renewcommand{\arraystretch}{1}
\begin{tabular}{c c c c c} 
     \textbf{Parameter} & \textbf{DP-SGD} & \textbf{DPKD} & \textbf{DP Syn} & \textbf{Ours}\\ [0.5ex] 
     \hline
     \multicolumn{5}{c}{Teacher}\\ [0.5ex]
     Epochs & - & 10\textsuperscript{\rm 1}, 20\textsuperscript{\rm 2, 3} & 10\textsuperscript{\rm 1}, 20\textsuperscript{\rm 2, 3} & 10\textsuperscript{\rm 1}, 20\textsuperscript{\rm 2, 3} \\ 
     Learning Rate & - & 1e-4\textsuperscript{\rm 1, 2, 3} & 1e-4\textsuperscript{\rm 1, 2, 3} & 1e-4\textsuperscript{\rm 1, 2, 3}\\
     Batch Size & - & 4096\textsuperscript{\rm 1, 2, 3} & 4096\textsuperscript{\rm 1, 2, 3} & 4096\textsuperscript{\rm 1, 2, 3}\\  
     Clipping Norm & - & 1.0\textsuperscript{\rm 1, 2, 3} & 1.0\textsuperscript{\rm 1, 2, 3} & 1.0\textsuperscript{\rm 1, 2}\\ 
     \hline
     \multicolumn{5}{c}{Student}\\ [0.5ex]
     Epochs & 25\textsuperscript{\rm 1}, 40\textsuperscript{\rm 2, 3} & 25\textsuperscript{\rm 1}, 40\textsuperscript{\rm 2, 3} & 5\textsuperscript{\rm 1}, 3\textsuperscript{\rm 2, 3} & 12\textsuperscript{\rm 1, 2, 3}\\ 
     Learning Rate & 1e-4\textsuperscript{\rm 1, 2, 3} & 1e-4\textsuperscript{\rm 1, 2, 3} & 1e-6\textsuperscript{\rm 1}, 2e-5\textsuperscript{\rm 2, 3} & 8e-5\textsuperscript{\rm 1, 2, 3}\\
     Batch Size & 4096\textsuperscript{\rm 1, 2, 3} & 4096\textsuperscript{\rm 1, 2, 3} & 16\textsuperscript{\rm 1, 2, 3} & 16\textsuperscript{\rm 1, 2, 3}\\  
     Clipping Norm & 1.0\textsuperscript{\rm 1, 2, 3} & 1.0\textsuperscript{\rm 1, 2, 3} & - & -\\ 
\end{tabular}
\caption{Training hyperparameters for the teacher and student model of every method for the \textsuperscript{\rm 1}Yelp, \textsuperscript{\rm 2}Big Patent, and \textsuperscript{\rm 3}DBpedia datasets.}
\label{table:hyperparameters}
\end{table*}
\section{Experiments}

\subsection{Experimental Setup}

We experimentally evaluated the privacy-utility tradeoff of our framework with the following setup. We used GPT2-Large \cite{radford2019language} and DistilGPT2 \cite{sanh2019distilbert} as our teacher and student model, respectively. GPT2-Large contains 774M parameters while DistilGPT2 contains 82M parameters, so the teacher model is about 9.5 times larger than the student. Our implementation of fine-tuning language models with DP and knowledge distillation utilizes the dp-transformers library \cite{inan2022dp-transformers}, which combines the NLP library Huggingface \cite{wolf2019huggingface} with the DP library Opacus \cite{yousefpour2021opacus}.

We experimented with the Yelp Open dataset \footnote{\url{https://www.yelp.com/dataset}}, which contains review text data on businesses. We follow the experimental setup as \citet{yue2022synthetic} by using the review stars and business category as the control codes. These control codes are constructed as "\textit{Business Type: Restaurant $|$  Review Stars: 3.0"} and are prepended to each sample. The top 10 frequent business categories are sampled and reviews with no ratings are removed. The resulting data split is $1.9$M reviews for training, $5000$ for validation, and $5000$ for testing. Each data sample is set to a sequence length of $128$.

We also experimented with the Big Patent dataset \cite{sharma2019bigpatent}, which consists of 1.3 million records of U.S. patent documents along with human-written abstractive summaries. For our experiments, we used the abstracts as the text data. To construct the control codes, we prepend each data sample with the Cooperative Patent Classification (CPC) code that it falls under. We only used a subset of the total number of documents, resulting in 150K abstracts for the training, 5000 for the validation, and 5000 for the testing dataset.

Lastly, we experimented on the DBpedia dataset \cite{zhang2015character}, which is an ontology classification dataset consisting of 14 non-overlapping classes from DBpedia 2014. Each of these 14 ontology classes are used as a control code for the training samples. In total, we randomly selected 195K training samples, 5000 samples for validation, and 5000 samples for the test set. 

We compare our method against three baselines: (1) a student model fine-tuned directly with DP-SGD, namely the teacher plays no role in training the student; (2) a student model fine-tuned using DPKD \cite{mireshghallah2022differentially}; (3) a student model fine-tuned only on DP synthetic data (DP Syn Data), i.e. $\lambda=0$ in Eq. \ref{eq:kd}, which follows the framework of \citet{yu2023training}. The training hyperparameters used for our method and the baselines with the teacher and student models are shown in table \ref{table:hyperparameters}.  We set the privacy budget to $\epsilon=2$, which is a strong privacy guarantee. We follow the standard practice of setting $\delta=1 / N$ where $N$ is the size of the dataset.

\begin{table*}[t!]
\begin{center}
\renewcommand{\arraystretch}{1.5}
\begin{tabular}{|c|c|c|c|c|c|c|} 
     \hline
     Model & Training & $\epsilon$ & Teacher & Yelp$\downarrow$ & Big Patent$\downarrow$ & DBpedia$\downarrow$\\  
     \hline
     GPT2-Large & DP-SGD & 1.0 & - & 26.84 & 21.91 & 28.85\\ 
     \hline
     DistilGPT2 & Zero-shot & 0 & - & 65.46 & 57.42 & 83.00\\ 
     \hline
     DistilGPT2 & DP-SGD & 2.0 & - & 48.12 & 41.80 & 60.81\\ 
     \hline
     DistilGPT2 & DPKD & 2.0 & GPT2-Large & 46.16 & 41.20 & 59.85\\ 
     \hline
     DistilGPT2 & DP Syn Data & 2.0 & GPT2-Large & 61.84 & 49.26 & 59.87\\ 
     \hline
     DistilGPT2 & DistilDP & 2.0 & GPT2-Large & \textbf{44.15} & \textbf{32.43} & \textbf{49.11}\\ 
     \hline
\end{tabular}
\caption{Comparison between the different models, the training method, the privacy budget $\epsilon$, the teacher model, and the perplexity score on the Yelp and Big Patent Dataset. Lower PPL means better utility.}
\label{table:yelp_results}
\end{center}
\end{table*}

For the generation of synthetic text, we use the truncation decoding strategies of top-$k$ \cite{fan2018hierarchical} and top-$p$ \cite{holtzman2019curious} sampling using $k=50$ and $p=0.9$. The teacher model generates 400K synthetic samples using control codes from the Yelp and DBpedia training dataset and 200K synthetic samples from Big Patent. The maximum sequence length of each generated sequence is $128$. We follow the methodology of \citet{kurakin2023harnessing} for training on DP synthetic data by splitting the Yelp and DBPedia synthetic samples into $396$K for the training set and $4$K for the validation. Likewise, $198$K and $2$K Big Patent synthetic examples for the training and validation set, respectively. For both our method and DPKD, we set $t=1.0$ and $\lambda=0.4$ for Equation \ref{eq:kd}. One minor technicality of our method is that because our teacher model was trained on control codes, its output distribution on the initial context vectors will be skewed toward the types of control codes. Hence, we ignore the first few softmax predictions in the distillation loss for each sample. We evaluated the utility of each method by measuring the perplexity (PPL) of the test set. 

\subsection{Main Results}\label{sec:main_results}

Table \ref{table:yelp_results} shows the results. We see that for the zero-shot performance of the student model, i.e. $\epsilon=0$, which is just a pre-trained DistilGPT2 model, it achieves a perplexity score of $65.46$, $57.42$, and $83.00$ on Yelp, Big Patent, and DBpedia, respectively. After performing DP-SGD with $\epsilon=2$, we observe that the student utility improves to $48.12$, $41.80$, and $60.81$, a significant utility improvement over the pre-trained student. However, compared to the DP fine-tuned GPT2-Large model with $\epsilon=1$, there is still much improvement to strive for. 

The DP-SGD baseline, where the student is privately fine-tuned without any teacher, serves as a lower bound for the utility of the student. This means that our framework must perform at least better than DP-SGD, or else the knowledge distilled from the teacher only hurts the student's utility.

For the DPKD baseline, we evenly allocated the privacy budget $\epsilon$ between the teacher and student, i.e., $T$ and $S$ use $\epsilon=1$ for DP-SGD. We observe that DPKD improves over the DP-SGD baseline by $2$, $0.6$ and $0.96$ PPL on the Yelp, Big Patent, DBpedia datasets, respectively. Hence, DPKD can give a modest performance boost to the student models compared to directly privately fine-tuning them for generative tasks, which was not the case for classification tasks shown in \citet{mireshghallah2022differentially}.

\begin{figure*}[t!]
\centering
\begin{subfigure}{0.475\linewidth}
\centering
\begin{tikzpicture}
\begin{axis}[
    width=0.8\linewidth,
    xlabel={$\lambda$ values},
    ylabel={PPL},
    xmin=0, xmax=1.0,
    ymin=30, ymax=70,
    xtick={0,0.2,0.4,0.6,0.8,1.0},
    ytick={40,50,60,70},
    legend pos=north west,
    ymajorgrids=true,
    grid style=dashed,
]

\addplot[
    color=blue,
    mark=square,
    ]
    coordinates {
    (0, 61.84)(0.4, 44.15)(0.5, 45.26)(0.6, 45.28)(1.0, 45.61) 
    };
    
\end{axis}
\end{tikzpicture}
\caption{Ablation study on distillation loss weighing parameter $\lambda$ for values between 0.0 and 1.0.}
\label{plt:albation_lambda}
\end{subfigure}
\hfill
\begin{subfigure}{0.475\linewidth}
\centering
\begin{tikzpicture}
\begin{axis}[
    width=0.8\linewidth,
    xlabel={$t$ values},
    ylabel={PPL},
    xmin=1.0, xmax=3.0,
    ymin=35, ymax=60,
    xtick={1.0, 2.0, 3.0},
    xticklabels={1.0, 2.0, 5.0},
    ytick={40,50,60},
    legend pos=north west,
    ymajorgrids=true,
    grid style=dashed,
]

\addplot[
    color=blue,
    mark=square,
    ]
    coordinates {
    (1.0, 43.99)(2.0, 47.44)(3.0, 56.33) 
    };
    
\end{axis}
\end{tikzpicture}
\caption{Ablation study on temperature $t$ for values between 1.0 and 5.0.}
\label{plt:albation_temp}
\end{subfigure}
\caption{Ablation study on $\lambda$ and $t$.}
\end{figure*}
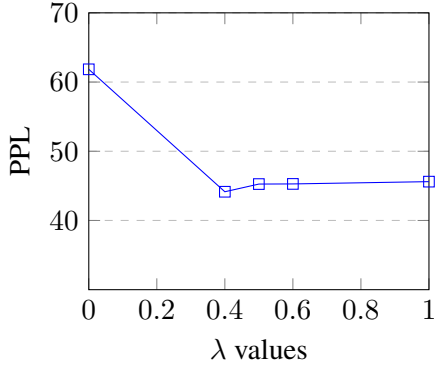
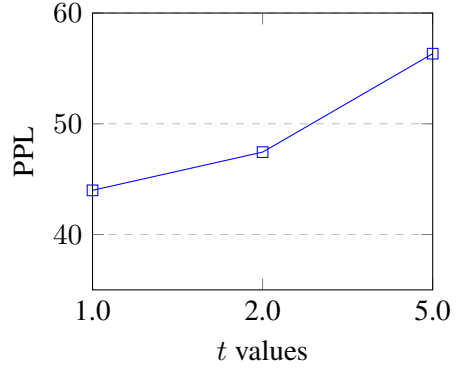

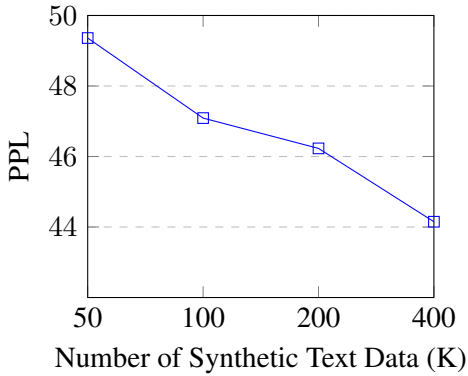
\begin{figure}
\centering
\centering
\begin{tikzpicture}
\begin{axis}[
    width=0.8\linewidth,
    xlabel={Number of Synthetic Text Data (K)},
    ylabel={PPL},
    xmin=1, xmax=4.0,
    ymin=42, ymax=50,
    xtick={1, 2, 3, 4},
    xticklabels={50, 100, 200, 400},
    ytick={44,46,48, 50},
    legend pos=north west,
    ymajorgrids=true,
    grid style=dashed,
]

\addplot[
    color=blue,
    mark=square,
    ]
    coordinates {
    (1, 49.36)(2, 47.09)(3, 46.23)(4, 44.15) 
    };
    
\end{axis}
\end{tikzpicture}
\caption{Ablation study on the number of Synthetic text data for values between 50K and 400K.}
\label{plt:albation_num_syn_data}
\end{figure}

DistilDP, using a GPT2-Large model with $\epsilon=2$ as the teacher, was able to substantially outperform all baselines for both datasets. For the Yelp datasets, our method beats all baselines by at least $2$ PPL. The margin is much larger for the Big Patent and DBpedia dataset, as our method outperforms all baselines by at least $9$ PPL. One contributing factor for the margin being larger for Big Patent and DBPedia is that the difference in size between the synthetic and original dataset is smaller than for Yelp. Most importantly, the results for both our method and DPKD demonstrate that the traditional knowledge distillation of aligning outputs between the teacher and the student can boost the performance of the student for differentially private generative tasks.

One insightful observation is that for the Yelp and Big Patent datasets, DP Syn Data severely lags behind all baselines, only providing a modest improvement over the zero-shot baseline but not improving over the DP-SGD baseline. However, we do observe that for DBpedia, DP Syn Data can match the utility of DPKD. Additionally, we found that it is harder to optimize DP Syn Data due to the student easily overfitting the synthetic review data. This result highlights the fact that training on DP synthetically generated data alone is insufficient to competitively train a compressed DP student model for generative tasks. We surmise that a teacher model at the scale of a large foundation model is needed to generate high enough quality synthetic text to overcome this limitation \cite{yu2023training}.

\subsection{Ablation Study}
In this section, we explore different hyperparameter values to study how they affect the performance of our framework using the Yelp dataset.

Figure \ref{plt:albation_lambda} plots the change of the PPL score as $\lambda$ changes. We see that relying purely on the distillation loss $\lambda=1$ already achieves competitive performance, capturing most of the knowledge from the private dataset. Meanwhile, the best utility we could achieve when fine-tuning only on the synthetic data, $\lambda=0$, is a severe utility degradation compared to $\lambda=1$. Interestingly, nearly evenly combining both losses, $\lambda = 0.4$, leads to the most optimal performance, necessitating both the supervised learning and distillation loss in Equation \ref{eq:kd}. 

Figure \ref{plt:albation_temp} illustrates the influence of the temperature parameter $t$ on the performance of our method. We explored values $t \in \{1.0, 2.0, 5.0\}$ and found that $t=1$ seemingly obtains the best utility while larger values of $t$ only hurt the performance. This is because as $t$ gets large, the output distribution becomes more uniform, which makes it harder for the student to learn the prediction uncertainty of the teacher. Additionally, the distillation loss takes over as it scales by $t^2$.

Lastly, Figure \ref{plt:albation_num_syn_data} shows how varying the number of synthetic text reviews affects the performance of the student. To get synthetic datasets of different sizes, we randomly sample subsets of the total 400K generated text dataset. We observe that the student greatly benefits from a larger training dataset, as more synthetic text data leads to lower PPL scores. Increasing the number of synthetic reviews from 50K to 400K improves the PPL of the student by 5. We suspect that by increasing the number of synthetic text data to more than 400K, we would observe our method outperforming all baselines by an even larger margin in the main results. We leave it as a future work to explore how large the synthetic text generation can be before we observe marginal gains in utility.

\subsection{Exploiting Hidden Representation of The Teacher}\label{sec:hid_rep}

The main results in section \ref{sec:main_results} showed that there is still a reasonable utility gap between the PPL score of the DP teacher model with $\epsilon=1$ and the DistilDP student model with $\epsilon=2$. The difference for the Yelp dataset is 17.3 perplexity, while the difference for Big Patent is less substantial, about 10.5 perplexity. In this section, we attempt to add more ways for the teacher to distill knowledge onto the student via a Mean Squared Error (MSE) loss $\alpha \mathcal{L}_{\text{MSE}}$ on the hidden representations to Eq \ref{eq:kd}. The rationale behind this method is that it would be beneficial to align the hidden representations of the student with the hidden representations of the teacher, assuming that the teacher has better hidden representations. Hence, we can use the MSE loss to push the student to mimic the hidden representation of the teacher. Note that since $T$ was fine-tuned using DP-SGD, by the post-processing property of DP, adding the MSE loss to Eq \ref{eq:kd} will not leak additional privacy.

\begin{table}[t!]
\begin{center}
\begin{tabular}{c c c c} 
     \hline
     Model & Training & Teacher & PPL$\downarrow$\\ 
     \hline
     GPT2 & DP-SGD & - & 31.41\\ 
     \hline
     DistilGPT2 & Ours & GPT2 & 37.17\\ 
     \hline
\end{tabular}
\caption{Further reducing the utility gap between the teacher and the student by adding an MSE loss to Eq \ref{eq:kd}. The results are for $\epsilon = 2$ and use the Big Patent dataset.}
\label{table:cosine_results}
\end{center}
\end{table}

\begin{table}[t!]
\begin{center}
\begin{tabular}{c c} 
     \hline
     Losses Used & PPL$\downarrow$\\ 
     \hline
     $\mathcal{L}_{\text{CE}} + \mathcal{L}_{\text{KL}} + \mathcal{L}_{\text{MSE}}$ & 37.17\\ 
     \hline
     $\mathcal{L}_{\text{CE}} + \mathcal{L}_{\text{KL}} + \varnothing$ & 37.42\\ 
     \hline
\end{tabular}
\caption{Measuring the impact of the MSE Loss $\mathcal{L}_{\text{MSE}}$ with combined with other losses.}
\label{table:cosine_impact}
\end{center}
\end{table}

To test the effect of aligning the hidden representation of the teacher and student, we applied the MSE loss to the last hidden representation. Since the dimensions of the last hidden representations must be equal between the teacher and the student, we use GPT2 as the teacher model, which has the same dimension as the DistilGPT2. We ran our experiments on the Big Patent dataset, and we set the MSE loss coefficient $\alpha=0.4$.

Table \ref{table:cosine_results} compares the performance between the teacher and the student. We see that the utility difference between the teacher and the student is even smaller, only 6 PPL. Table \ref{table:cosine_impact} shows that the inclusion of the MSE loss into Eq \ref{eq:kd} does mildly improve the overall utility of the student. We suspect that the distillation loss $\mathcal{L}_{\text{KL}}$ already captures most of the information distilled from the teacher. However, the improvement from the MSE loss suggests that our framework can be further improved by including more information about the hidden representation of the teacher. We leave it as a future work to explore other advanced knowledge distillation techniques that exploit this information.


\section{Conclusion}
In this work, we introduced DistilDP, a novel DP knowledge distillation algorithm by utilizing DP synthetic data generation. We presented comprehensive experimental results that demonstrate DistilDP obtains the best utility over all existing baselines for strict privacy settings. More generally, we showed that differentially private knowledge distillation for autoregressive large language models can be effective. The provided ablation study highlights that both the synthetic text as the hard label and the output distribution of the teacher as the soft label are crucial to boosting the performance of the student. We believe further improvements to our framework are feasible by generating more synthetic text data and including additional knowledge distillation from the hidden representations. 


\section{Limitations}
One limitation of our work is that the teacher model needs to be trained with DP-SGD. This limitation is because the teacher is a function of the private dataset, and is used for synthesizing text and aligning its outputs with the output of the student. However, the teacher is generally much larger than the student, and thus the computational and memory costs of running DP-SGD for the teacher are much larger than for the student. Furthermore, the problem setup of DP compression only requires differential privacy for the student, not for the teacher. A more optimal solution to this problem would only use DP-SGD for the student, and avoid applying DP-SGD for the teacher. However, such a solution is non-trivial to derive and is left as future work.

Another limitation is the same limitations from the synthetic text generation framework we used from \cite{yue2022synthetic} carry over to our framework. In particular, small classes corresponding to control codes are disproportionately affected, where tight DP guarantees most negatively impact learning the distributions of small-size classes. 

Finally, our work builds on recent advances in DP and LLM by initializing the models with pre-trained weights, and then privately fine-tuning them on a downstream dataset \cite{li2021large, yu2021differentially}. However, recent work has shown that public LLMs can leak private information \cite{carlini2021extracting}. A stronger and interesting future work would look to relax this assumption.

\section{Ethical Considerations}
In this work, we utilized pre-trained LLMs and well-known language modeling datasets that were accessed from the Hugging Face API, which are publicly available and free to use. The Big Patent dataset is licensed under CC BY 4.0. The GPT2- models are licensed under the Apache License, Version 2.0. We could not find a license for the Yelp dataset, however, the intended use of this dataset must be for academic research, which we do follow. Our intended use of the artifacts in this work is academic in nature, which is aligned with the intended use of the creators. We did not anonymize the dataset since Big Patent does not contain personally identifiable information, and we removed attributes that contain any private information, such as user or business id, from the Yelp dataset, as the dataset we worked with only contained text data and non-private categorical data, which are used as control codes. Additionally, we did not disclose the contents of either dataset. Our use of public models and datasets minimizes any unintended privacy leakage that could result from experimenting. 

\section*{Acknowledgements}
This work is supported by the Defense Advanced Research Projects Agency (DARPA) under Contract Nos. HR001120C0088, NSF award number 2224319 and DGE-1842487, REAL@USC-Meta center, and gifts from VMware. The views, opinions, and/or findings expressed are those of the author(s) and should not be interpreted as representing the official views or policies of the Department of Defense or the U.S. Government.

\bibliography{main}

\appendix
\newpage
\section{Additional Experimental Setup}
\label{sec:appendix}

Most of the hyperparameter values from table \ref{table:hyperparameters} were selected from \cite{yue2022synthetic} and \cite{mireshghallah2022differentially}, due to the large computational cost of hyperparameter searching of DP-SGD. Such values, like the clipping norm, are standard values to use. However, certain values were selected via hyperparameter sweeping, such as the learning rate. The computing infrastructure used to train the teacher and student models with DP-SGD is 8 40GB A100 GPUs, and the students with DP synthetic text data used 8 Quadro RTX 5000. The total training time for the teacher models on the Yelp dataset was 2 days and 17 hours, while on the Big Patent dataset, it took 10 hours. For the student models, it took about one days for the DP-SGD baseline, one and a half days for the DPKD baseline, and about five hours for DP Syn Data and ours. The reported main results are based on one run of training and then inference on the testing set. The version we used for the dp-transformers library \cite{inan2022dp-transformers} is 1.0.1 which was installed by source. 

\end{document}